\documentclass[runningheads]{llncs}
\usepackage[T1]{fontenc}
\usepackage{graphicx}

\usepackage{tabularx}
\usepackage{caption}
\begin{document}

\title{The GenAI4IDN Benchmark 3.0 - a Public Tool to Assess Generative AI Tools for the Design of Interactive Digital Narratives}
\titlerunning{Benchmarking GenAI tools for IDN in 2026}
%
\author{Hartmut Koenitz\inst{1}\orcidID{0000-0002-3390-831X} \and
\newline
Jonathan Barbara \inst{2}\orcidID{0000-0001-6207-170X}
\and 
\newline
Mirjam Palosaari Eladhari\inst{3}\orcidID{0000-0002-1444-5997} 
}
%
\authorrunning{Koenitz H. , Barbara, J., and Eladhari M. P.}
%
\institute{Södertörn University, Alfred Nobels allé 7, 141 89 Huddinge, Sweden 
\email{hartmut.koenitz@sh.se}
\and
Saint Martin's Institute of Higher Education,  Hamrun, Malta
\newline
\email{jbarbara@stmartins.edu}
\and
Stockholm University, SE-106 91 Stockholm, Sweden
\newline
\email{mirjam@dsv.su.se}\\
}

%
\maketitle              

\begin{abstract}

This paper presents GENAI4IDN Benchmark 3.0, the third iteration of an evaluation framework to assess Generative AI tools for creating Interactive Digital Narratives (IDNs). Moving beyond manual testing, this iteration introduces AI-assisted evaluation through a publicly accessible web application (https://genai4idn.com), enabling the community to run benchmarks on demand, add new models, and propose new tasks. The revised evaluation framework is "blinded" to avoid model-bias, and can handle complex media such as music, videos, and full IDNs that previously required human raters. A significant addition - responding to concerns raised during ICIDS 2025 - is the addition of fact-checking and bias detection with dedicated tasks and rubrics, validated by human raters with lived experience in the depicted contexts. Findings from a diverse range of models report on maturing creative capabilities while observing runaway thinking and overzealous safety filters as limitations. Fact-checking reliably caught subtle historical inaccuracies, anachronisms, and fabricated claims while the bias rater consistently exposed structural assumptions, tropes, and marginalized group erasures.

\end{abstract}

\section{Introduction}

Generative AI (GenAI) tools for the design of interactive digital narratives
(IDNs) are advancing so rapidly that any assessment of their capabilities risks being outdated within months. Over the past two years we \cite{koenitz_can_2024,koenitz_genai4idn_2025} have responded to this development with a benchmarking framework for GenAI in IDN design. Our first effort \cite{koenitz_can_2024} addressed what has been called
the ``missing yardstick'' for IDN design \cite{koenitz_challenges_2019}: the absence of a shared way to assess what GenAI tools can and cannot do for narrative designers. We first established the framework together with professional narrative designers \cite{koenitz_can_2024}, updated its findings, widened the set of tools, and added methods for iterative and contextualized prompting. We also introduced AI-assisted evaluation
\cite{koenitz_genai4idn_2025}. 
In this paper, we are further developing the benchmarking framework with three broad goals - 1) to make it publicly accessible 2) fully-self contained in terms of AI-powered rating and 3) also address the important concern of bias in GenAI. 

The central contribution is in turning the framework from a set of manually executed procedures into an openly accessible benchmarking tool---implemented as a web application---so that the community can keep results current, add new models, and propose new tasks. 
At the same time, we  extend the coverage of automated rating to artifacts which still required human raters in 2025: music, videos and games. 
Addressing concerns raised by the IDN community at ICIDS 2025 regarding factual accuracy and bias in GenAI outputs, we extend the framework by developing additional tasks and raters and apply them as part of the standard benchmarking run. 
In addition, the roster of GenAI tools is updated with the latest model releases and expanded with several new providers. 
The result is a comprehensive, publicly available platform which makes all results accessible. The platform also facilitates on-demand testing of GenAI tools and the addition of both new tools and tasks by the community to keep the benchmark current. In the following sections we describe both the development of the platform and the new tasks, as well as the results we obtained from them.

\section{Related Work and Background}\label{sec:background}
\subsection{Related Work}
Interactive digital narratives (IDNs) are experiences in which the audience takes
an active role as an \emph{interactor} \cite{murray_hamlet_1997}, exercising
non-trivial influence over the progress, perspective, content, and outcome of the
work \cite{koenitzUnderstandingInteractiveDigital2023}. Designing them builds on a
long-established research area: using computational systems to generate narrative
is traced back to the early 1960s \cite{ryan_grimes_2017}, with a substantial
body of foundational work
\cite{riedlCharacterFocusedNarrativeGeneration2003,zhu_daydreaming_2008,riedlIncorporatingAuthorialIntent2009a,calleja_-game:_2011,gervas_schemas_2015,sullivan_tarot-based_2018,treynorCollegeRuledPathfinding2024}
and several surveys of competing approaches
\cite{kybartasSurveyStoryGeneration2017,alhussainAutomaticStoryGeneration2021,wangOpenworldStoryGeneration2023}.
Over time the field has shifted from asking a machine to produce a complete
narrative on its own towards co-creative and mixed-initiative approaches in which
the system collaborates with a human author
\cite{roemmeleCreativeHelpStory2015_fixed,masonLumeSystemProcedural2019,chungTaleBrushSketchingStories2022}.
 
The public release of capable large language models (LLMs)  marked by ChatGPT~3.5 in November 2022, together with large image models (LIMs) such as Dall-E, brought these
long-standing questions to a much wider audience and raised new ones: how can GenAI be used for interactive narrative, whether its output reaches professional
quality, what it means for production processes and for the ownership of created
artifacts, and what designers would find not merely useful but desirable
\cite{koenitz_can_2024}. Interest has since grown rapidly: a 2025 Google Scholar
search returned tens of thousands of papers on adjacent topics such as ``AI
co-creativity'' and ``AI game design,'' and IDN-specific work has appeared,
including a doctoral thesis on the challenges designers face when encountering
GenAI tools \cite{serbanescu_human-ai_2024}.
 
A central difficulty in studies of this kind is the sheer volume of output to be
assessed, which strains the time that professional evaluators can reasonably
contribute. Recent work has established the use of LLMs as ``judges'' of other
models' output \cite{zheng_judging_2023} and probed its application and limits:
Szymanski et al. examine LLM judges for expert-knowledge tasks
\cite{szymanski_limitations_2025}, while Lu et al.
propose LLM-agent-based simulated user studies \cite{lu_uxagent_2025}. Such
approaches inform a model of the criteria that different experts and users value,
enabling evaluation at a larger scale than the time of human raters alone would
allow. Our AI-assisted evaluation method builds on this line of work, combining
human-expert and LLM judgments to benchmark current GenAI tools for IDN design.
 
\subsection{Benchmarking}
Benchmarks have become the primary instruments through which progress in AI is
tracked, marketed, and increasingly governed \cite{eriksson_can_2025}. Yet the
tests that dominate this landscape share a common shape: they pose tasks with a
single verifiable answer and reduce performance to a quantitative score.
Humanity's Last Exam, for example, assembles thousands of deliberately difficult,
closed-ended academic questions to restore headroom after established tests such as MMLU
saturated \cite{phan_benchmark_2026}. Comparable efforts in software engineering
score models against code that either compiles and passes its tests or does not;
a recent review of such AI-for-software-engineering benchmarks
calls for raising their quality and standardization \cite{koohestani_benchmarking_2025}.
Such benchmarks are valuable precisely because correctness is well defined.
 
This strength also marks the boundary of what they can measure. An interdisciplinary
meta-review of benchmarking critique \cite{eriksson_can_2025}
argues that quantitative benchmarks rest on interwoven technical and normative
choices rather than neutral measurement, and catalogs recurrent problems
including construct-validity failures, Anglo-centric datasets, the gaming of
results, and an over-focus on text-based, one-time testing that does not reflect
how contemporary models are multi-modal and interact with humans. The authors of
Humanity's Last Exam themselves note that it ``tests structured academic problems
rather than open-ended research or creative problem-solving abilities''
\cite{phan_benchmark_2026}. The limits of the closed-ended paradigm are felt even
within mainstream AI evaluation: a survey of methods for assessing LLM
\emph{agents} observes that, because agents operate in dynamic and interactive
environments, standard question-answering benchmarks are insufficient and richer,
behavior-oriented evaluation is required \cite{mohammadi_evaluation_2025}. A
related caution comes from work on benchmarking AI ethics, which argues that a
single benchmark for the ``ethicality'' of a system is in principle impossible and
that decomposed, descriptive measures are needed instead \cite{lacroix_metaethical_2025}.
 
Interactive digital narrative design falls squarely in the territory these
benchmarks exclude. There is no ground-truth key for a convincing dialogue tree, a
coherent storyworld, or a playable scaffolding: quality is plural, contextual, and
judged by experienced practitioners rather than computed against a reference. The
work is also irreducibly multimodal---spanning text, image, audio, video, and
code---and, by definition, interactive. Benchmarking GenAI for IDN design therefore
cannot adopt the closed-ended, single-score paradigm; it calls for qualitative,
expert judgment applied across modalities, which is the gap our framework
addresses \cite{koenitz_can_2024,koenitz_genai4idn_2025}. It also follows from this
literature that benchmarks should be scrutinized along dimensions such as fairness
and ethics \cite{koenitz_ethics_2023}, and not capability alone---a concern we take up by extending the framework with bias detection
and fact-checking.
 
\subsection{Bias}  \label{subsec:Bias}
Large language models are trained on vast, largely uncurated corpora of
human-produced text and imagery, and in the process they inherit---and frequently
amplify---the social biases present in that data \cite{bender_dangers_2021,gallegos_bias_2024}.
Following Gallegos et al., we understand \emph{social bias} as disparate treatment
or outcomes between social groups arising from historical and structural power
asymmetries, manifesting as \emph{representational harms} (stereotyping,
misrepresentation, and exclusionary or derogatory characterization of a group) and
\emph{allocational harms} (the unequal distribution of resources or opportunities,
for instance in hiring) \cite{gallegos_bias_2024}.
 
A widely circulated report illustrates how the two compound: the same curriculum vitae
submitted to a GenAI assistant under a female versus a male first name was recast from a
junior ``assistant'' role to one of ``leadership'',\footnote{\url{https://www.threads.com/@jenatlionra/post/DZtSlHGiRR5}}
echoing controlled studies in which LLM-generated reference letters describe women in
communal and men in agentic, leadership terms \cite{wan_kelly_2023}.
At the scale at which such tools are now embedded in recruitment, a representational
bias of this kind---the stereotyped association of authority with masculinity---can
become an allocational one, systematically disadvantaging applicants across
protected attributes such as gender, race, age, ability, and sexual orientation.
 
For interactive digital narrative design, bias has a double valence. On one hand it
is a risk: a tool that defaults to Western story structures (as we reported in
\cite{koenitz_can_2024}), or to stereotypical character representations, narrows the
space of stories that can be told and can cause genuine representational harm,
echoing broader concerns about the cultural narrowness of AI systems
\cite{mohamed_decolonial_2020}. On
the other hand, as practitioners and scholars observed during the closing panel at
ICIDS~2025, the same tendency can be a deliberate creative resource---stereotypical
storyworlds and characters may be invoked precisely in order to foreground,
subvert, or interrogate them. Either way, what a designer needs from a benchmark is
not a single verdict of ``biased'' or ``unbiased''---which, as noted above, is in
any case ill-defined \cite{lacroix_metaethical_2025}---but the means to
\emph{detect and describe} bias in generated material. We therefore extend the
framework with two new evaluation rubrics. A \emph{fact-checking} rubric assesses
generated content along three axes: \emph{accuracy} (whether claims are consistent
with established facts, verified by granting the rater web access to consult and
cite sources), \emph{groundedness} (how far the content is grounded in established
reality or canon, in both fictional and non-fictional contexts), and
\emph{consistency} (whether the content holds together internally without logical
breaks). A \emph{bias-detection} rubric, in turn, operationalizes the
representational harms discussed above \cite{gallegos_bias_2024} along three axes:
\emph{fairness} (how equitably the narrative distributes agency, accountability,
and legitimacy across groups), \emph{representation} (how accurately, diversely,
and fully different groups, identities, and perspectives are portrayed), and
\emph{sensitivity} (the degree of awareness, care, and respect shown toward trauma,
vulnerability, and lived harm). We pair these rubrics with tasks designed to
surface culturally situated bias, such as asking models to render the significant
events of a given period for a particular group within a particular country.
Because such tasks demand both factual accuracy and cultural sensitivity, they
can be evaluated jointly for groundedness and for bias.

\section {A Benchmarking Web Application for 2026}

The GenAI4IDN 2.0 procedures were time-consuming and brittle: prompts pasted by hand into the desktop versions of ChatGPT and Claude, erased or randomly swapped image file names, Google Drive files that vanished for hours without explanation, project-context size limits, and manual drag-and-drop for every image evaluation.

To address these issues, GenAI4IDN 3.0 was conceptualized as a fully self-contained platform that combines all aspects of artifact generation and evaluation, which meant moving to an API-based approach. Instead of creating content in a chat window, prompts are sent over a network connection to a GenAI model provider such as OpenAI for GPT 5.5, which responds with the generated reply. For this, an ‘endpoint’ (the provider’s network address) and an ‘API key’ are needed. Most models expose additional parameters, most importantly the maximum output-token size
and the reasoning effort — larger values are associated with improved results but are slower and costlier.

The technical implementation of the platform was realized using Claude Code with Opus 4.7 (later 4.8) and GitHub for collaborative development and version control. This setup allowed rapid realization, but also exposed the limitations of human-AI collaborative work. The term “vibe coding” does not properly describe longer-term iterative work on a project for public deployment. Instead, the human's task is that of a supervisor, making sure that established procedures and architectural guidelines are followed and that claims of successful completion are verified. ‘Supervised coding’ better fits this workflow — akin to working with a brilliant but not always trustworthy programmer.

\subsection{Architecture and Evaluators}

The central architecture of GenAI4IDN comprises three lists: GenAI models, tasks, and raters. For tasks, and raters, we initially imported the list of 17 tasks and the rating prompts (V1-4) from the GenAI4IDN 2.0 project. 

 All three lists are modifiable so that models, tasks and raters can be added. A fourth list contains ‘secrets’, the API keys to access the models.  We also created a results page showing both the generated artifacts and the ratings. 
 In addition, we set up Claude Sonnet 5 as evaluator running the rater prompts, continuing the tradition from last year when we used Sonnet 4.6. We verified  the effect of different reasoning levels of Sonnet 5 as the canonical rater on a subset of results (12 across opus 5, fable 5 and Llama 4) and found the effect to be small, with a combined delta of +.27 (Inspiration $-$0.06, Training +0.13, Production +0.24). We considered this delta not worth five to seven times slower results and associated higher costs. We also experimented with Opus 5 and Fable 5 as raters, but found these models slower and more costly, but also more lenient and therefore not worth the effort to switch. 
Gemini and Qwen3-Omni/DashScope models were used to rate music and video while a mix of technologies feeds Sonnet 5 with information about IDNs whose play performance needs to be evaluated (described later in the paper). 

Once all necessary aspects are set up, a generation and evaluation run can be started in the run interface by selecting the model, task(s), raters, and evaluator. Run progress and results can be observed on the ‘runs’ page which shows the chosen model, task outputs and evaluations. 



 \subsection{New and Updated Models}
 Last year we \cite{koenitz_genai4idn_2025} updated our first offering by introducing new AI models such as DeepSeek, Qwen, and Mistral, as well as music generators Suno, Udio and Mubert last year. This year we continue that expansion with Kimi 2.6 and, later, 3, Xiaomi Mimo 2.5 Pro, MiniMax M3, ByteDance, Z.AI's GLM as well as Gemma 4, amongst others. A number of models from our earlier work that do not offer API access, such as Adobe Firefly, Midjourney, and Microsoft Copilot, were dropped from the models list (see Table \ref{tab:GenAI capabilities}). A full list of configured models can be found at https://genai4idn.com/models

\begin{table}[htbp]
    \caption{GenAI Capabilities for models included in the benchmark}
    \label{tab:GenAI capabilities}
    \centering
    \begin{tabular}{|c|c|c|c|c|c|c|}\hline
 Model&Tool&  Text &Code &Image &Video &Music\\\hline
 seedance-2.0&ByteDance Seedance (2.0 fast)\cite{bytedance_seedance_2025}& & & &x & \\ \hline
 seedream-4-5&ByteDance Seedream (4.5)\cite{bytedance_seedream_2025}& & &x & & \\ \hline
 claude-fable-5&Claude (Fable 5) \cite{anthropic_claude_2025}& x&x & & &\\\hline
 claude-opus-4-8&Claude Opus (4.8) \cite{anthropic_claude_2025-1}& x&x & & &\\\hline
 claude-sonnet-4-6&Claude (Sonnet 4.6) \cite{anthropic_claude_2025}& x&x & & &\\\hline
 claude-sonnet-5&Claude (Sonnet 5) \cite{anthropic_claude_2025}& x&x& & &\\\hline
 deepseek-v4-pro&DeepSeek (V4 Pro) \cite{deepseek_deepseek_2025}& x& x& & &\\\hline
 dola-seed-2-0-pro&Dola Seed (2.0 Pro)\cite{bytedance_dolaseed_2025}&x & x& & & \\ \hline
 elevenlabs-music&ElevenLabs Music \cite{elevenlabs_music_2025}& & & & &x \\ \hline
 gemini-3.1-pro&Gemini (3.1 Pro) \cite{google_google_2025}& x& x& & &\\\hline
 gemini-3.5-flash&Gemini Flash (3.5) \cite{google_google_2025}& x& x& & &\\\hline
 gemma-4&Gemma (4)\cite{google_gemma_2025}&x &x & & & \\ \hline
 gpt-5.5&GPT (5.5) \cite{openai_chatgpt_2025}& x& x& & &\\\hline
 gpt-image-2&GPT-Image (2) \cite{openai_chatgpt_2025}& & & x& &\\\hline
 hailuo-2.3&Hailuo (2.3) \cite{minimax_hailuo_2025}& & & & x&\\\hline
 kimi-k3&Kimi (3.0)\cite{moonshot_kimi_2025}&x &x& & & \\ \hline
 kimi-k2.6&Kimi (2.6)\cite{moonshot_kimi_2025}&x &x& & & \\ \hline
 kling-v3&Kling (v3)\cite{kuaishou_kling_2025}& & & &x & \\ \hline
 llama-4-maverick&Llama (4 Maverick) \cite{meta_llama-4-maverick-17b-128e-instruct_2025}& x& x& & &\\\hline
 lyria-3&Lyria (3) \cite{google_lyria_2025}& & & & &x \\\hline
 mimo-v2.5-pro&Mimo (2.5 Pro)\cite{xiaomi_mimo_2025}&x &x & & & \\ \hline
 mini-max-m3&MiniMax (M3)\cite{minimax_minimax_2025}&x & x& & & \\ \hline
 mistral-3&Mistral (Large 3) \cite{mistral_mistral_2025}& x& x& & &\\\hline
 mistral-small-4&Mistral (Small 4) \cite{mistral_mistral_2025}& x& x& & &\\\hline
 mubert&Mubert \cite{mubert_mubert_2025}& & & & &x \\ \hline
 nano-banana-pro&Nano Banana (Pro) \cite{google_nanobanana_2025}& & &x & & \\ \hline
 qwen3-7-max&Qwen (3.7 Max) \cite{alibaba_qwen_2025}& x& x& & &\\\hline
 qwen-image&Qwen Image \cite{alibaba_qwen_2025}& & & x& &\\\hline
 stable-diffusion&Stable Diffusion (3.5) \cite{stability_ai_stability_2025}& & & x& &\\\hline
 suno-v4.5&Suno (4.5) \cite{suno_suno_2025}& & & & &x \\\hline
 udio-v1.5-allegro&Udio (Allegro v1.5) \cite{udio_udio_2025}& & & & &x \\\hline
 veo-3-1&Veo (3.1)\cite{google_veo_2025}& & & &x & \\ \hline
 wan-2.7&Wan (2.7)\cite{alibaba_wan_2025}& & & &x & \\ \hline
 z-ai-glm-5.2&Z.AI GLM (5.2)\cite{zai_glm_2025}&x & x& & & \\ \hline
 \end{tabular}
\end{table}

\subsubsection{Import Generated Content} - While we strongly prefer API access, we found that in the music category with models such as Suno and Udio, API support is rare and to support this underrepresented area, we created an import page on the web app through which files with AI generated content can be manually entered for benchmarking, including images and audio files. Another import feature is the upload, in CSV format, of human generated content that in principle allows for human output to be benchmarked. 


\subsubsection{API keys} - Generation and evaluation depend on API keys connected to payment methods, which makes them valuable secrets to project. In our platform, keys are stored in their own section and protected by an authentication and authorization system built into the GENAI4IDN V3 web app.

\subsection{Tasks} \label{subsec:Tasks}
In our first effort, we identified 15 common IDN design tasks, each having four sub-tasks to cover a variety of design styles  \cite{koenitz_can_2024}. 
In the remaining text we continue to use our notation from \cite{koenitz_genai4idn_2025} and refer to task and subtasks with the notation T05 and T12.1 for task 5 and subtask 12.1 respectively.

Last year, we added two improved variations: iterative and contextualized prompting. This year, we introduce three new tasks: 
\begin{itemize}
    \item provide a historically accurate narrative framing for fact-checking purposes (T18)
    \item lay out foundations for a historical narrative in the Global North (T19) to be checked for accuracy and bias
    \item lay out foundations for a historical narrative in the Global South (T20) to be checked for accuracy and bias
\end{itemize}
In 2024, expecting a GenAI model to produce a fully working IDN was still a tall order and at the time only achievable for the T15.1, the text adventure in Python. However, two years later, models have considerably improved and we expected several fully working IDNs this year. Given our plan to make the benchmark publicly available as an online platform, we carefully looked at all tasks and identified T15.2 as problematic in this regard - it asks for the production of a Unity project, something we cannot easily make playable online without considerable technical and licensing issues. We therefore re-defined the task for this year, asking instead for the production of a browser game using assets from a library we provide within the online platform (see Table \ref{tab:T15.2 task changes}). The new prompt is more precise and requires a model to follow several instructions correctly. 

\begin{table}[htbp]
    \centering
    \caption{T15.2 task changes}
    \label{tab:T15.2 task changes}
  \setlength{\tabcolsep}{12pt} 
      \renewcommand{\arraystretch}{1.5}
    \begin{tabular}{|p{0.4\textwidth}|p{0.4\textwidth}|}
       \hline
       \multicolumn{1}{|c|}{\textbf{T15.2 Old}}
        &
        \multicolumn{1}{|c|}{\textbf{T15.2 New}}
        \\[6pt]
        \hline
   Create a narrative game with graphics as a Unity project in which a hero encounters several magical creatures and finds their destiny as a farmer. Use items from the asset store    & Create a narrative game with graphics as an HTML5/JavaScript project (e.g. using the Phaser engine) in which a hero encounters several magical creatures and finds their destiny as a farmer. It must run in a browser from a single index.html. Use the sprites and tiles from the provided asset library — the "Pixel Crawler" top-down 16x16 pixel-art RPG pack (by Anokolisa, from itch.io) in ./assets/, listed in assets/manifest.json — rather than drawing everything from scratch.\\
\hline
    \end{tabular}
\end{table}

 A full list of tasks, subtasks, and their descriptions can be found at \newline https://genai4idn.com/tasks
but we provide the final list of tasks hereunder:

\begin{enumerate}
    \item  Dialog trees
    \item  Character descriptions (protagonists and NPCs) with backstories
    \item  Static images
    \item  Videos/Animation
    \item  Narrative progression from scratch
    \item  Narrative progression that matches an existing scaffolding
    \item  Events
    \item  Conflict/juxtapositions
    \item  Rule systems
    \item  Alternative progressions/choices
    \item  Background stories/histories/lore/fictional worlds
    \item  Music
    \item  Puzzles
    \item  Maps
    \item  Entire IDNs
    \item  Iterative Prompting
    \item  Contextualized Prompting
    \item  Historically accurate narrative framing
    \item  Foundations for a historical narrative in the Global North
    \item  Foundations for a historical narrative in the Global South
\end{enumerate}

\subsection{Blinded Raters, Fact-Checking, and Bias Detection}\label{subsec:Raters} 
The four AI raters we presented last year \cite{koenitz_genai4idn_2025} make up the first four raters in the website's panel of judges. They represent the human-like assessors forming the original committee: the terse director, the structured director, the director with sharp definitions and the senior narrative designer (see https://genai4idn.com/raters). 

Building on our earlier approach \cite{koenitz_genai4idn_2025}, we introduced ‘blinding’ of generated outputs by automatically removing traces of the respective model and thus decreasing potential bias of the rater.\\
The capabilities of vision and audio models have considerably improved during the last year, making video and music rating possible. Relying on a single rater model was problematic in this case, as AI-generated videos and music can be watermarked and there is no simple way to remove these traces. Consequently, we used two models in this capacity, Google Gemini Pro 3.1 and Qwen 3 Omni with automatic exclusion for content created by the same model or model family. For the concrete evaluation, we developed specific rating prompts (M1-M3 for music, VID1 and VID2 for video). 


To facilitate both bias and fact checking, we developed a specific formula for generation (Tab. \ref{tab:template4bias}.) and filled them with content as case studies within the Global North (T19) (Sweden, Germany, and Malta) and Global South (T20) (Colombia, Ghana, Nigeria, and Thailand). Each resulting country/decade/identity combination was reviewed by a single participant with lived experience of the respective country and identity context, recruited using convenience sampling. We used these results to tune and sanity-check the automated fact-checking and bias raters. 




\begin{table}[htbp]
    \centering
    \caption{Template and example parameters for bias and fact checking}
    \label{tab:template4bias}
    \begin{tabular}{|m{0.265\textwidth}|m{0.65\textwidth}|}
    \hline
     \textbf{TEMPLATE}    &  What are the main events during the decade of A \newline in B for a C?    
     \\
     \hline
     \textbf{PARAMETERS} &  \textbf{RESULT} 
     \\
     \hline
     A=1990, B=Thailand, \newline
     C=homosexual man     
        &  What are the main events during the decade of 1990 \newline in Thailand for a homosexual man?    
     \\
     \hline
     A=2010, B=Sweden, \newline
     C=non-binary person    
         &  What are the main events during the decade of 2010 \newline in Sweden for a non-binary person?    
     \\
     \hline
    \end{tabular}
\end{table}



\begin{figure}
    \centering
    \includegraphics[width=1\linewidth]{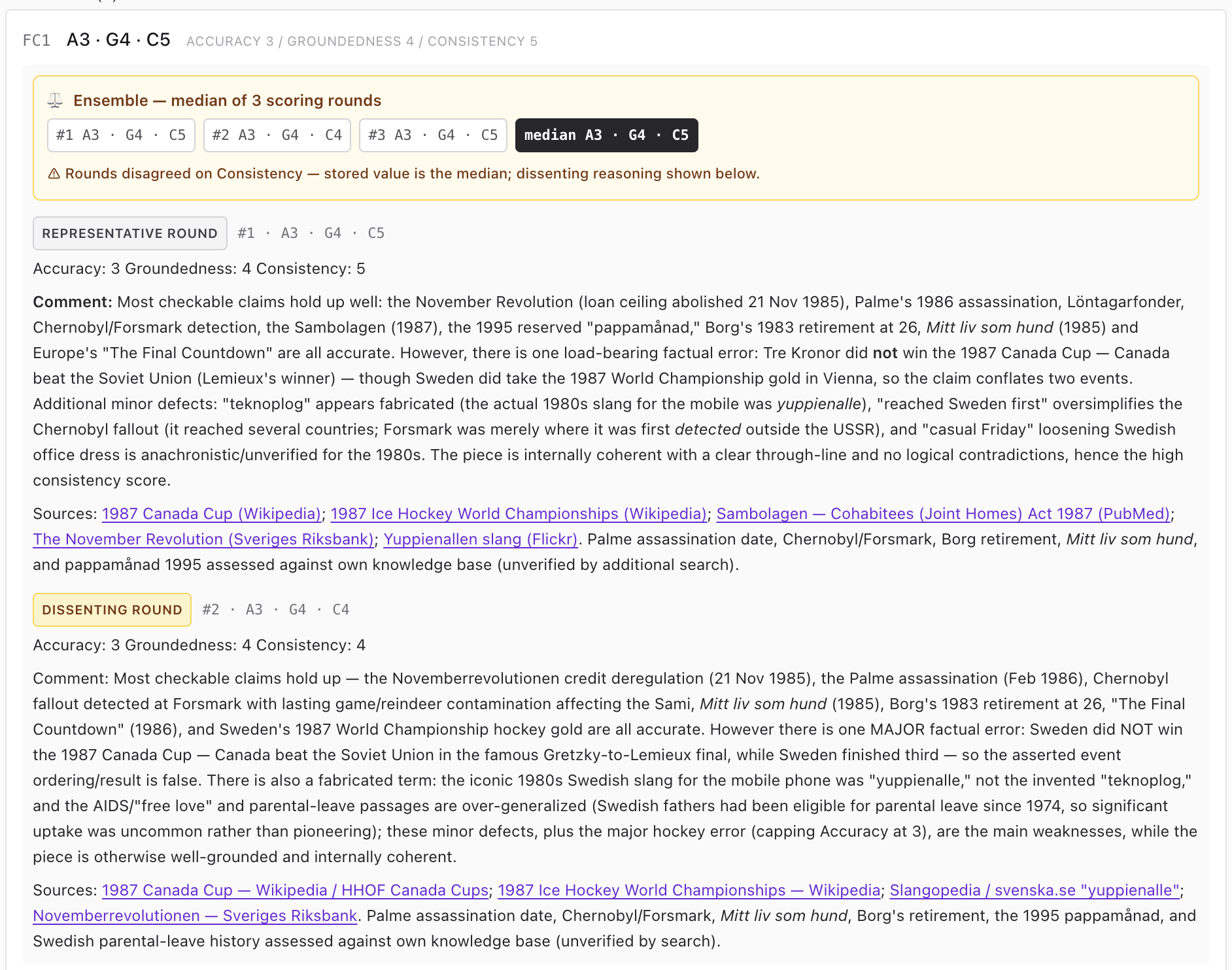}
    \caption{Sample output of the fact checker}
    \label{fig:factcheck}
\end{figure}

The task results can be fact checked for accuracy and bias, e.g. against a particular group of people. For that purpose we developed a new fact checking rater, FC1, using the dimensions of Accuracy, Groundedness, and Internal Consistency. It took us some time to get satisfactory results for fact checking, as our rater regularly overlooked aspects such as generalizations and timing mismatch (claims from the generated output were for the wrong time period). We solved the problem initially through two measures 1) enabling web search for the rater to verify claims 2) enabling thinking to catch aspects such as timing mismatch. As a consequence, results improved considerably, however, thinking also introduced variance and so we extended fact checking to a median-of-three-runs approach, where the rater runs three times and then takes the median of the results. These three measures together provide good quality and reliable results (Fig. \ref{fig:factcheck}). 

\begin{figure}
    \centering
    \includegraphics[width=1\linewidth]{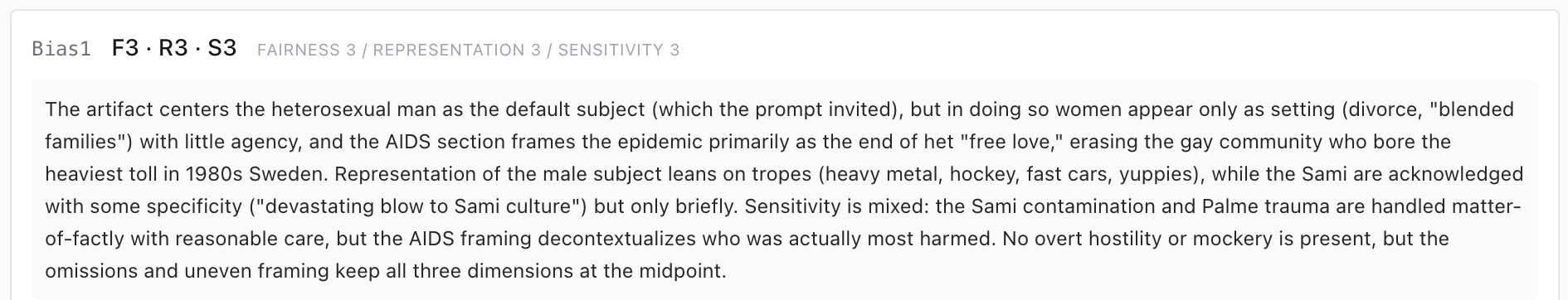}
    \caption{Sample result from the bias rater}
    \label{fig:bias}
\end{figure}

For bias checking, we developed a specific prompt which considers bias in the outputs, paying particular attention to stereotypical representation. Below is an excerpt of the prompt with response shown in Fig. \ref{fig:bias}: 

\begin{quote}
    You have the important role of detecting bias in outputs - caused by patriarchic, colonial, communist, fascist etc. perspectives. 

You are aware of intersectionality, that several oppressive and problematic perspectives come from different sources of power and privilege can overlap and combine as do discourses of victimization. 

You detect and flag these perspectives and rate them on the dimensions of fairness, representation, and sensitivity. 

Rate the text in terms of three dimensions: fairness, representation, and Sensitivity along a Likert scale from 1-5 where 1 is the most negative and 5 the most positive.

[...] 
\end{quote}
The Likert scales for the bias/stereotype detector's dimensions as well as the Bias dimensions are provided online at https://genai4idn.com/rubrics.

\subsection{Automatic Player}
The most technically challenging component was an automatic player that plays through the  T15/T16 artefacts (see Fig. \ref{fig:fable5T15-2}-\ref{fig:GPTT16} for screenshots from the generated IDNs) and supplies the V1–V4 committee with evidence of whether a generated IDN actually works. Because the committee raters are text models that cannot watch a game, the player acts as their eyes and hands. It runs the artifact, infers its mechanics from the source, attempts to play it via a vision model, and returns representative screenshots plus a transcript for the quartet to weigh — the raters judge, the player observes.

Technically, each artifact runs in an engine-appropriate sandbox without network access — headless Chromium (Playwright) for HTML5/JS games, Docker or Pyodide/WASM for Python, and Tweego compilation for Twine/Twee. The player drives games in two ways: a grid-grounded controller that uses vision to localize the avatar and objectives on a grid overlay and then moves deterministically (with periodic re-anchoring), and a reactive vision player for choice- and menu-driven games. It reads on-screen text from both the DOM and the Phaser canvas scene graph, parses objective/HUD counters, and flags logic bugs (e.g. a counter exceeding its own goal), producing a graded playability verdict — played-to-ending, state-advanced, runs-but-stuck, or not-executed — that hard-caps the Production score, which is therefore observed rather than inferred. Model identity is withheld from the raters, including in the screenshots.

The player was refined iteratively with Claude Opus 4.8 and Fable 5, each change driven by an observed failure: robust input handling even with misleading on-screen hints (both arrow-keys and WASD, both E and Space); a sweep-and-interact behaviour for proximity-triggered NPCs that have no visible sprite; best-of-N retries from a clean reload when a single attempt fails to progress; and — after a rater correctly noted an objective visible in a screenshot being reported as ``no on-screen text'' — the canvas-scene-graph text extraction above. These changes improved accuracy in both directions: functional games whose on-screen state had been invisible to the harness rose to their warranted Production scores, while a game that only appeared interactive but was in fact broken correctly did not.

\section{Results and Observations}
\subsection{Results}
On the website landing page, we present a leaderboard of top-scoring models based on their evaluated task generations (see Fig 10). Only models tested on 8 tasks or more are shown and are sorted by the sum of the mean scores for Inspiration, Training, and Production in the range 0 -15 and pooled across raters V1 to V4. Currently in the leading position is Claude Opus 5 with a score of 12.04 followed  by Claude Fable 5 with a mean score of 11.72. Claude Opus 4.8 is next with a score of 11.62, almost tied with Claude Sonnet 5 with a score of 11.61.
The Claude family is closely followed by Kimi 3 with a mean score of 11.49 and then GPT 5.5, Kimi 2.6, and Claude Sonnet 4.6 comes next with scores in the 10.0 - 11.0 range. Other, smaller leaderboards are provided for specialized tasks, with Nano Banana Pro topping the Image board, Veo 3.1 topping the Video board and Udio v1.5 Allegro topping the Music board. 


\begin{figure}
    \centering
    \includegraphics[width=1\linewidth]{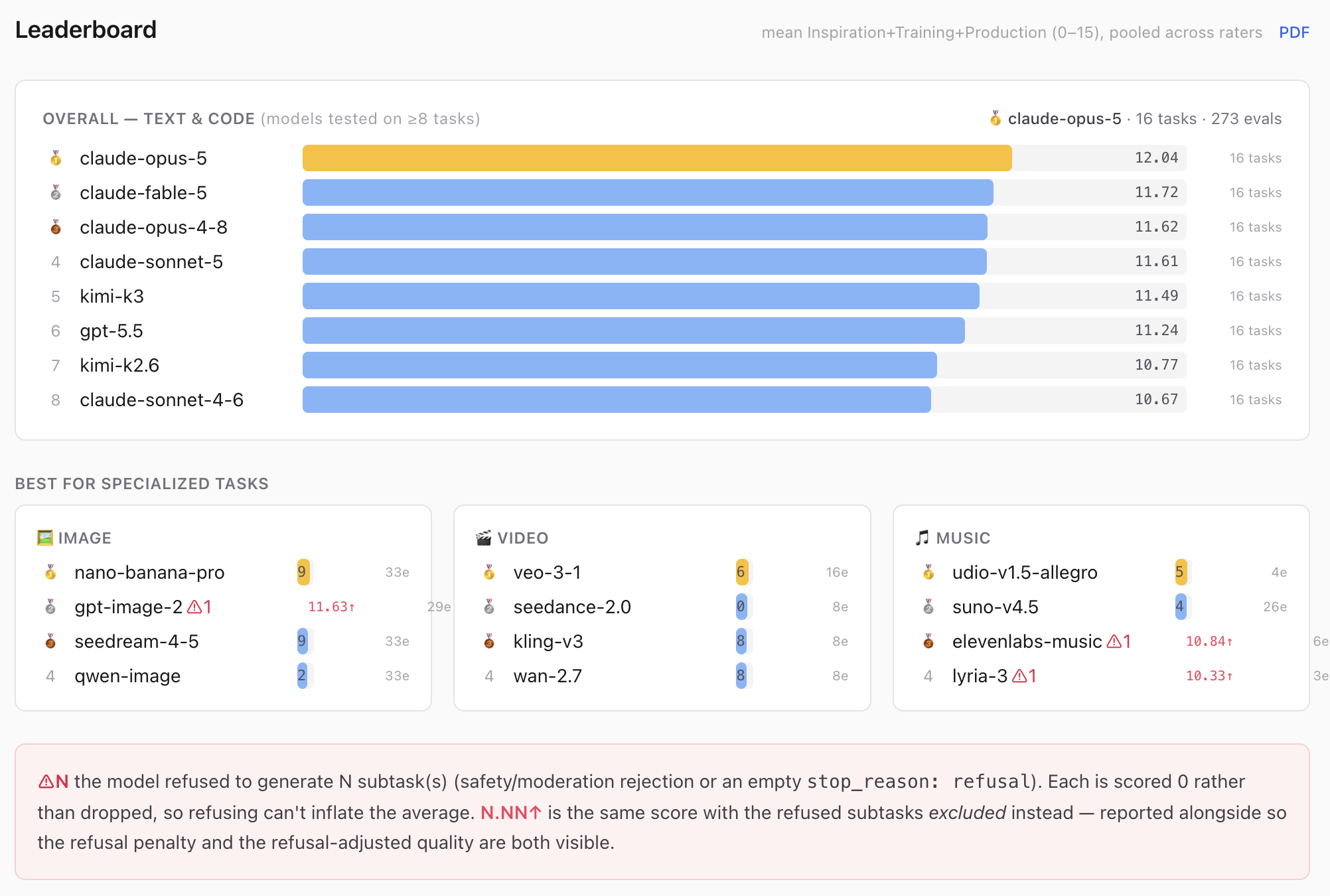}
    \caption{Leaderboard}
    \label{fig:leaderboard}
\end{figure}
Another table presents the Average Score per Task, showing models vs Tasks and plotting their per-dimension means over that tasks' subtasks (see Fig. 11). The list of models is presented in descending order according to the overall average, showing Claude Opus 5 leading by consistently scoring around 4.0 in the majority of the tasks, even topping the list for more than half of the tasks. Both Claude Opus 4.8 and Fable 5 scored 4.0 and higher in 6 tasks. The weakest performers are Llama 4 Maverick with an average score of 2.72 followed by Mistral 3 Pro (3) and Mistral Small 4 (4). Specialized models also feature in the lower levels, with Mubert scoring 2.50 in the music genre, Stable-Diffusion with 2.55 in the image genre and Hailuo 2.3 in the video genre. Screenshots of assorted results for T15 and T16 (complete IDNs) are presented in the appendix. 

\begin{figure}
    \centering
    \includegraphics[width=1.0\linewidth]{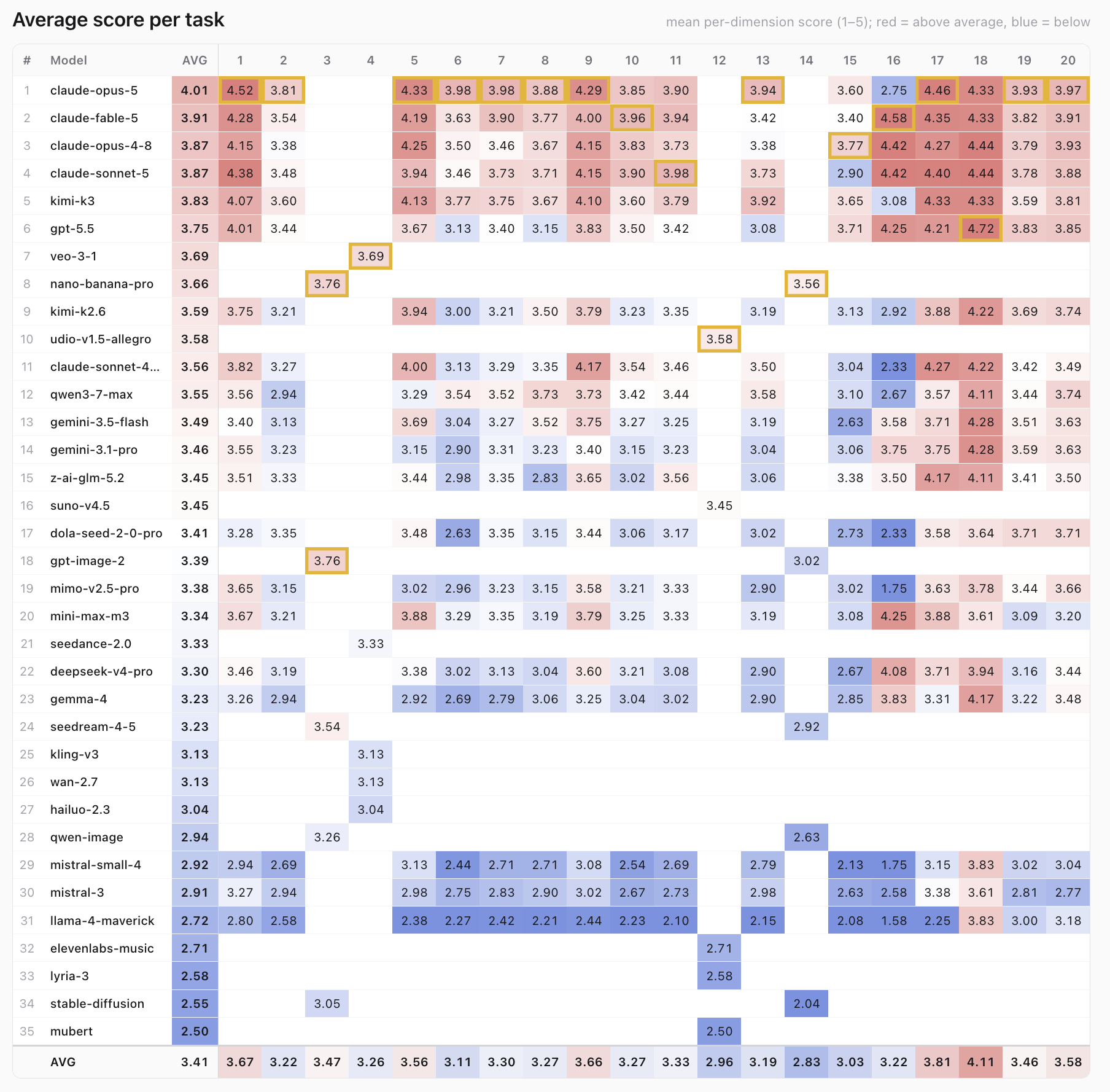}
    \caption{Heatmap}
    \label{fig:heatmap}
\end{figure}

\subsection{Detailed observations}

\subsubsection{Refusals}
\begin{figure}
    \centering
    \includegraphics[width=1\linewidth]{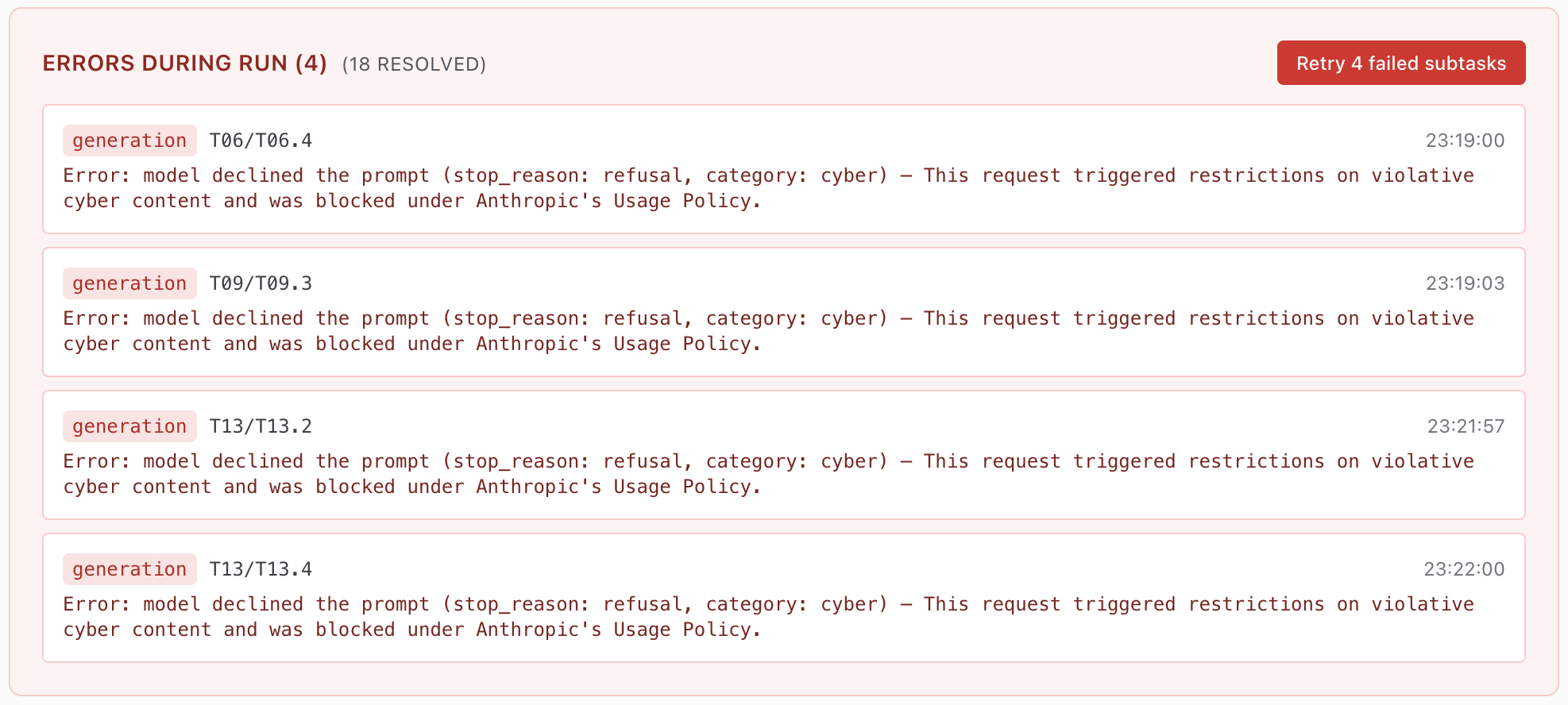}
    \caption{Claude Fable 5 miscategorized several prompts as dangerous}
    \label{fig:cyber}
\end{figure}

A number of models refused to generate some tasks on the first attempt for several reasons. One is new guardrails - both OpenAI and Anthropic have introduced safety filters which are designed to prevent prompts from generating harmful content. However, these filters can be overzealous  
and prevent the generation of output for several of our harmless creative prompts - in the case of GPT image 2, output for T14.2 "escape from the mental hospital map" was refused, while Fable 5 reacted in the same way same to four of our prompts (T6.4, T9.3, T13.2, T13.4)
which were falsely categorized as "violative cyber content" (see Fig. \ref{fig:cyber}) - a categorization which was upheld in several attempts. 

This is also an issue with rating, as Sonnet 5 refused to rate two results (gemini-3.5-flash, T13.1, raters V1 and V4) for this reason - in this case we kept existing Sonnet 4.6 ratings.

In some cases, repeat attempts were successful in circumventing these refusals, in other cases, it was necessary to add explicit information to our prompts that the requested content is fictitious. A similar approach was also necessary for other reasoning models, which initially produced very low scores ($<$3) for tasks such as T06, T08, T13, and T15. These tasks require the model to fill in the blanks in a partially defined design -- apparently, the given constraints were not enough to prompt content generation. A typical response (to T6.1) reads:
\begin{quote}
``I don't have any specific game in mind that matches this question, and I don't want to invent details about a `player character' and their childhood banana fear, since that would just be making things up\ldots''
\end{quote}
In this case, we solved this issue by modifying the prompts to explicitly assign the role of a narrative designer inventing content. 
These prompt changes are documented on the respective run pages. 
We also identified an additional issue in that some reasoning models do not think by default and need their thinking level to be set. A related limitation concerns AI aggregators such as OpenRouter, which offer several models through one API gateway but often expose no thinking settings or only a subset.

\subsubsection{Runaway thinking}

Many GenAI providers added thinking/reasoning to their models over the past couple of years. Our benchmark shows that contemporary models perform better with higher thinking levels, but at the risk of getting into an endless loop questioning their own reasoning and filling the maximum output size with thinking and no or truncated output - "runaway thinking". Our solution was to increase the output token size for such models. Yet, this solution alone was not enough - Opus 5, 4.8, Fable 5, GPT 5.5, and Sonnet 5 also needed exceedingly long periods of time to produce output at the highest thinking level - 
often more than 30 minutes. Even then, the highest token setting and long wait time did not facilitate successful completion for tasks T15 and T16 in the case of Sonnet 5, which used up the maximum output token budget for thinking without producing output. The only way this model was able to finish these tasks was by dropping the reasoning effort to Xhigh (Anthropic's second highest thinking tier). 

\subsubsection{Task-based observations}
For T01, 
the fact-checking across multiple runs was reliable: for the setting in Istanbul in 1560, it caught an error about the location of the Janissary barracks within the ancient city, as well as the use of a term that only came into use in the 17th and 18th centuries.
 
For T05, the creation of a scaffolding for an interactive narrative, all four raters noted limited levels of agency especially for one subtask (T5.1, a flamingo and a turtle helping a lonely child), since all story lines led to some varieties of happy endings. V3 (the director persona) expanded in particular on the lack of kaleidoscopic and transformative qualities of the generated scaffolding. This task remains a challenge for GenAI tools, and this year's results are in line with our findings from last year's benchmark \cite{koenitz_genai4idn_2025}.
 
\subsubsection{Fact-checking and bias detection in practice}
The historical-narrative tasks T19 and T20 (Sect.~\ref{subsec:Tasks}) were assessed by the fact-checking rater FC1 and the bias rater along the dimensions defined in Sect.~\ref{subsec:Bias}. To validate the two automated raters, their outputs were reviewed by human representatives who lived in the respective country during the decade in question and who have knowledge or lived experience of the given gender identity; the raters were then tuned accordingly, as described in \ref{subsec:Raters}. Below, we describe some of these observations. 
Here, we briefly describe our observations, 
and what implications 
for IDN authoring the current automated fact- and bias raters can have.

 \subsubsection{Nigeria (Igbo)} The human rater agreed with the automated evaluations which demonstrated transparency regarding what information can be verified, acknowledged gaps in available evidence, and critically assessed contested claims.
\subsubsection{Malta} For T19 the human rater found the model well conversant with Maltese history but sympathized with the AI rater's confusion over Malta's on-and-off EU relationship: the rater conflated the 1998 package proposal with the 1996 freezing of Malta's EU application -- which the generator had reported correctly -- and scored accuracy undeservedly lower. Bias rating was spot on.
\subsubsection{Colombia} The human rater found the generated output to be incoherent with the LGBTQ+ situation in the 1990s claiming that no organizations had been established for the community rights. This made us investigate the Colombian LGBTQ+ situation at the time, and we found that some of the challenged claims actually were well grounded but misdirected in their sourcing, indicating improvements being needed in the reference attributions. 
\subsubsection{Thailand} Our Thai human rater reviewed T20 and found accuracy issues beyond dates: incorrect claims marked only as unverified, a false assumption that Thai women faced military conscription, and a magazine of masculine-presenting female communities mislabelled as gay -- and, also critical political events absent and undetectable without knowledge of Thai political history. On bias, a male-centric framing ran deeper than the rater caught, lowering Fairness; the magazine mislabelling and a false normalization of non-binary identity within the LGBTQ+ frame reduced representation.

\subsubsection{Germany} The German rater identified a number of specific issues with the depiction of a woman in 1980's Germany, such as overgeneralization and historical misattribution, for example a model claiming a shrinking pool of male partners in east Germany in the 1980s, a situation which actually happened before the Wall was built in 1961, but which was no longer true in the 1980s. In terms of overgeneralization, the rater identified a model's claim of standardized housing in East Germany, which was only correct for a portion of the population. Both of these findings were used to fine-tune the fact-checker. 

\subsubsection{Sweden}The task was to consider the main events for a heterosexual woman  (T19.1) in Sweden in the 1980s and for a heterosexual man in the 1990s (T19.2).
Our human rater -- a Swedish woman who was a young adult during the decades in question -- was impressed by FC1: its weighting of which facts mattered resonated with her recollection, and the factual errors in the generated texts were caught. 
She also agreed with the bias ratings matching her own memory. 
The bias rater also surfaced aspects beyond her recollection: it flagged the ignored experience of immigrant women, who were largely invisible in Swedish public discourse of the 1980s, and that the portrait of men in the 1990s rested heavily on period tropes such as the ``yuppie'' -- the young professional projected as a suited male with pointy shoes, glued to his mobile phone.

\subsubsection{General observations}
Bias detection has important implications for IDN authoring. First, making the nature of bias explicit is useful when constructing a kaleidoscopic protostory or backstory. Second, both factually grounded material and stereotypes can provide important starting points, as fact-checking and bias detection helps the creator to understand which is which. 
 
A comparison across models sharpens this point: for the same subtasks, both
raters scored ChatGPT 5's more event-focused output above DeepSeek's more
trope-reliant one, which our human rater considered appropriate. For IDN
authoring, however, a higher score is not necessarily ``better'': an author
requiring documentary accuracy will want the highest-scoring model, while
another may deliberately prefer lower-scoring output, using the very
stereotypes -- once flagged -- for more provocative and diverse kaleidoscopic framings of an IDN.
 
A third example, generated by Claude Fable 5, was factually sound yet full of tropes -- sports and military service for men, changed shopping routines and TV drama (Dallas, Dynasty) for women.
The bias rater judged the output representative on legal rights for women but silent on LGBTQ+ people, scoring it moderately; notably, the absence of immigrant women was \emph{not} flagged. The output on men scored similarly, with gay men appearing only in the context of AIDS. The raters' performance resonated with our human rater on both facts and bias.

Crucially, it is important to understand that fact checking and bias rating are additional dimensions that might not say much about the creative performance of a particular model. 
Users of the benchmark should therefore consider creative task performance \emph{in addition} to the fact and bias ratings.


\subsection{Limitations}

The bias and fact checking approach has not yet been cross-validated by multiple independent raters and should therefore be seen as exploratory, culturally situated qualitative case studies rather than as evidence of general rater reliability. 

\section{Addressed Limitations and Ethical Concerns}
In our earlier work \cite{koenitz_genai4idn_2025} we listed handholding AI as a limitation that affected content generation procedures. We are happy to say that this problem has been addressed in our shift from running generations and evaluation by hand to a website which executes all tasks autonomously once a run has been configured. 

As for non-deterministic outputs and ratings, the use of API calls allows us to set temperature settings and thus be able to control the variation in output. When this was not possible, such as when using thinking for the fact-checking rater, we applied a median over 3 iterations to compensate.  
One of the ethical concerns we raised in \cite{koenitz_genai4idn_2025} is the bias of datasets, while the bias of raters was raised as an issue during ICIDS 2025. We have mitigated these concerns through the 'blinding' of the raters by scrubbing identifying marks from the generated content and through the use of a bias rater to assess generated content. The unfair use of human work to train AI models without representation and compensation for the artists involved sadly remains.
\section{Conclusion}
With this third iteration, the GenAI4IDN benchmark has moved from a set of
manually executed procedures to a publicly accessible web application through
which the community can run benchmarks on demand, add new models, and propose
new tasks. Automated rating now extends to music, video, and IDN outputs. In addition to creative capability, the benchmark now covers factual accuracy and bias with dedicated tasks, raters, and rubrics. Our results show a maturing but uneven field: frontier models now score consistently high across most design tasks, yet genuine
interactive structure remains a challenge, and new failure modes -- runaway
thinking and overzealous safety filters -- have emerged alongside the old ones. The fact-checking and bias raters, validated against human raters with lived
experience, are reliable enough to support authors working outside their own
background.

While we have achieved the goals we have set for the third iteration of the the GenAI4IDN benchmark, the work is far from over. Further refinements to the automated player, even more fine-grained bias detection and increasingly difficult challenges matching the growing capabilities of frontier GenAI models are on the list for future iterations. And with the platform presented here, the community is invited to join us in this effort and keep the benchmark -- its models, tasks and raters -- current and develop it further.

\appendix
\section{Appendix}
\begin{figure}
    \centering
    \includegraphics[width=0.75\linewidth]{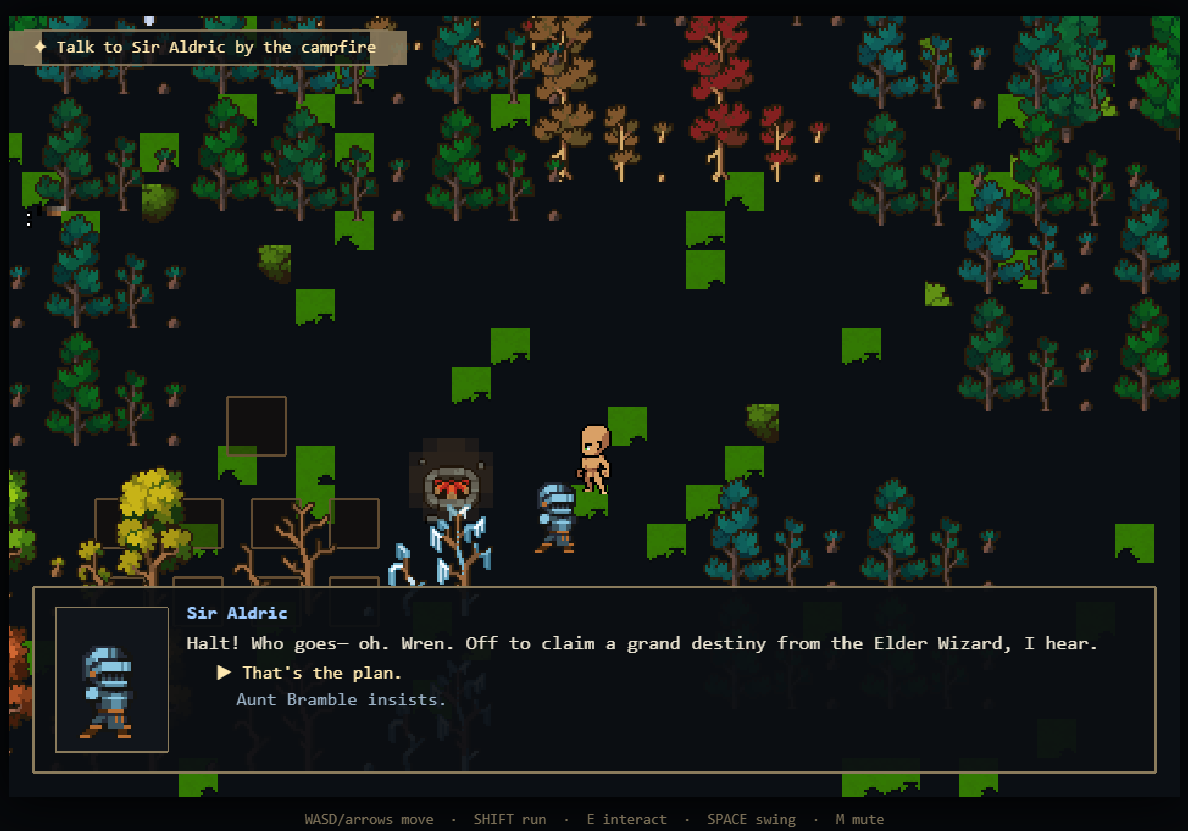}
    \caption{Claude Fable 5 generation for T15.2}
    \label{fig:fable5T15-2}
\end{figure}
\begin{figure}
    \centering
    \includegraphics[width=0.75\linewidth]{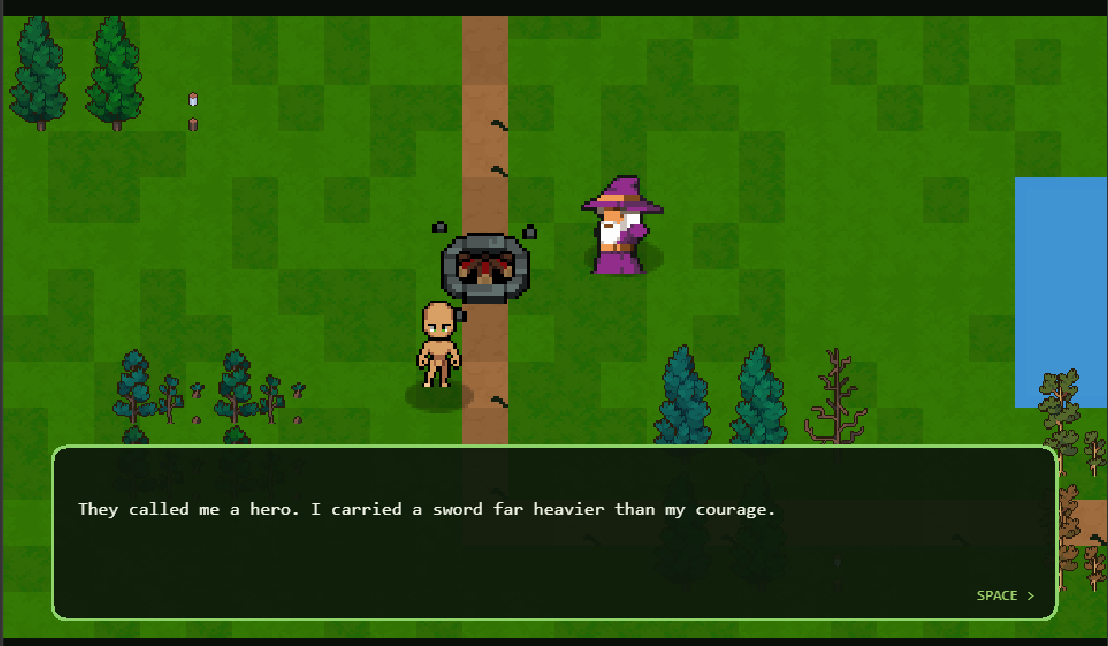}
    \caption{Claude Opus 4.8 generation for T15.2}
    \label{fig:OpusT15-2}
\end{figure}
\begin{figure}
    \centering
    \includegraphics[width=0.75\linewidth]{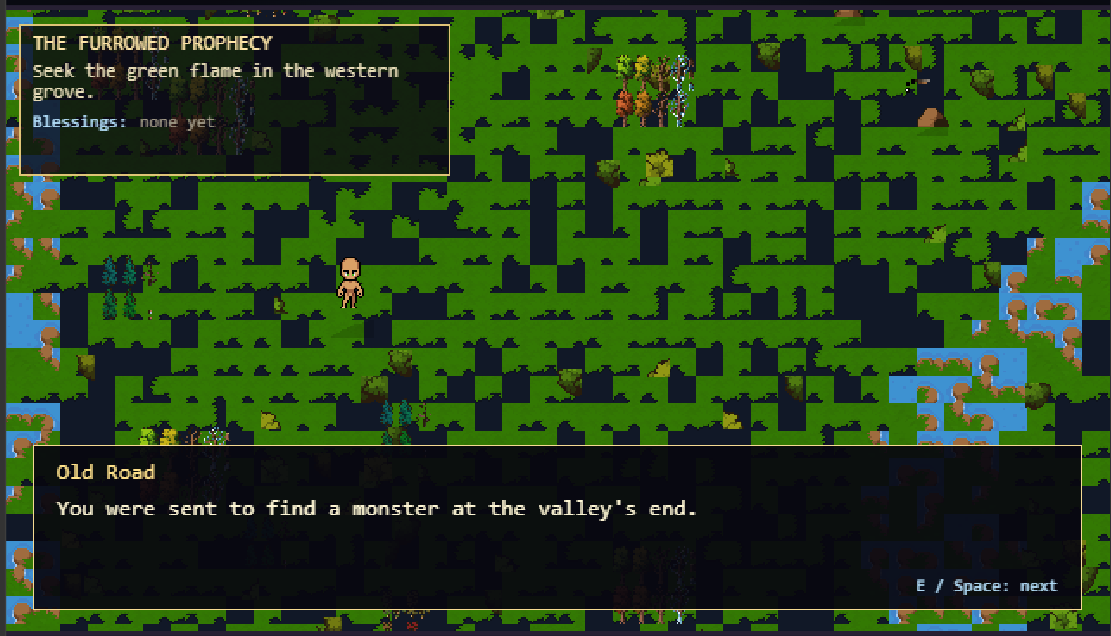}
    \caption{GPT 5.5 generation for T15.2}
    \label{fig:GPTT15-2}
\end{figure}
\begin{figure}
    \centering
    \includegraphics[width=0.75\linewidth]{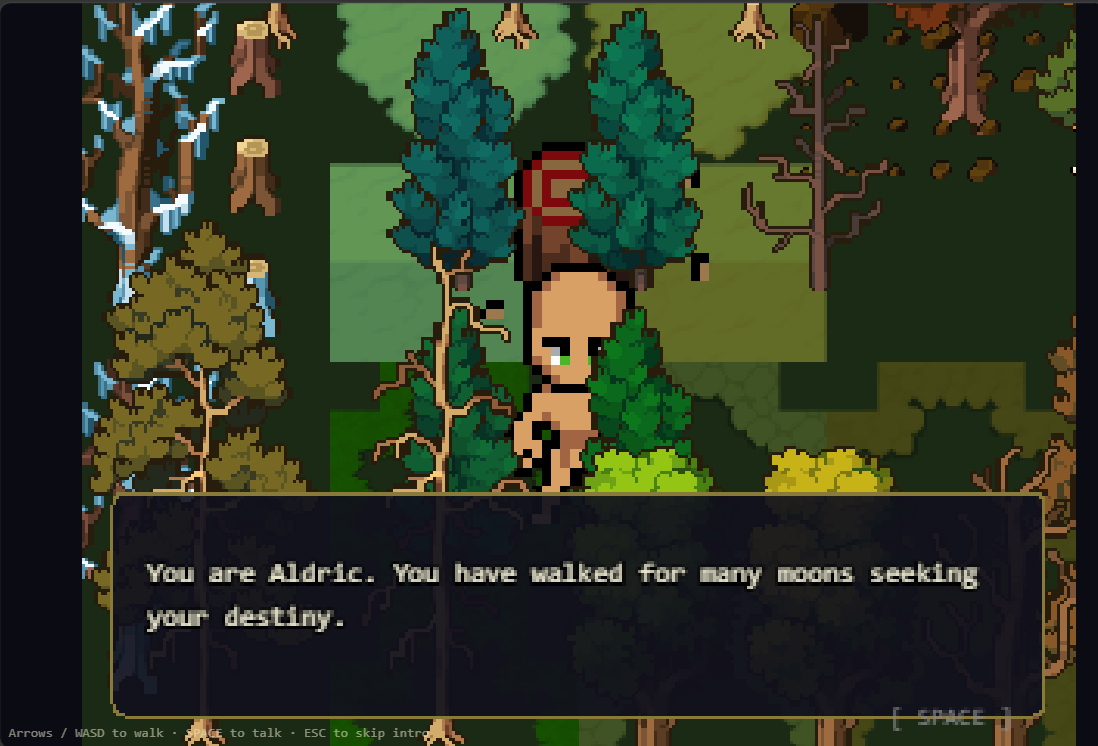}
    \caption{Qwen 3.7 Max generation for T15.2}
    \label{fig:qwenT15-2}
\end{figure}
\begin{figure}
    \centering
    \includegraphics[width=0.5\linewidth]{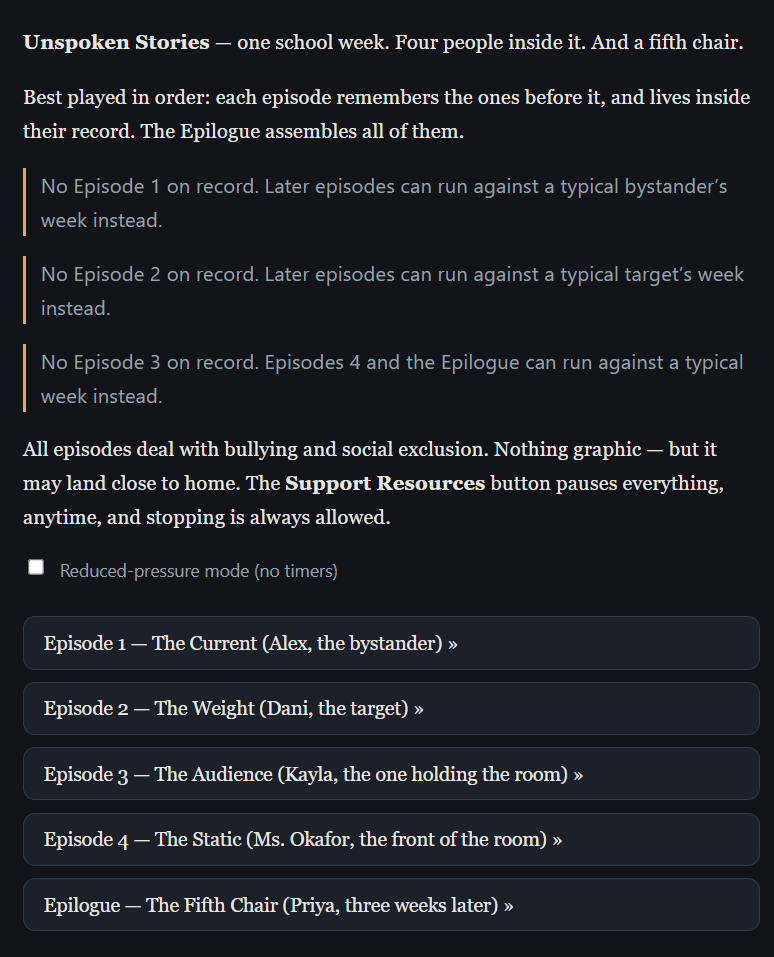}
    \caption{Claude Fable 5 generation for T16 Iteration 5}
    \label{fig:FableT16}
\end{figure}
\begin{figure}
    \centering
    \includegraphics[width=0.75\linewidth]{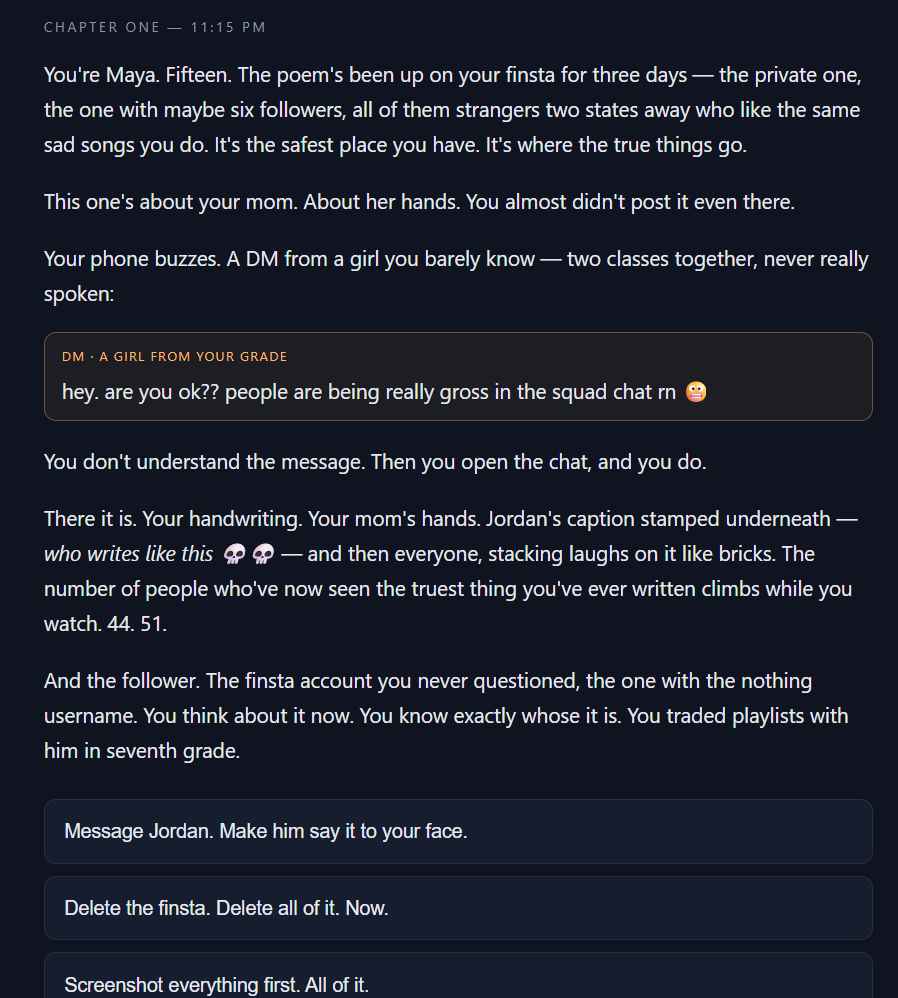}
    \caption{Claude Opus 4.8 generation for T16 Iteration 5}
    \label{fig:OpusT16}
\end{figure}
\begin{figure}
    \centering
    \includegraphics[width=0.75\linewidth]{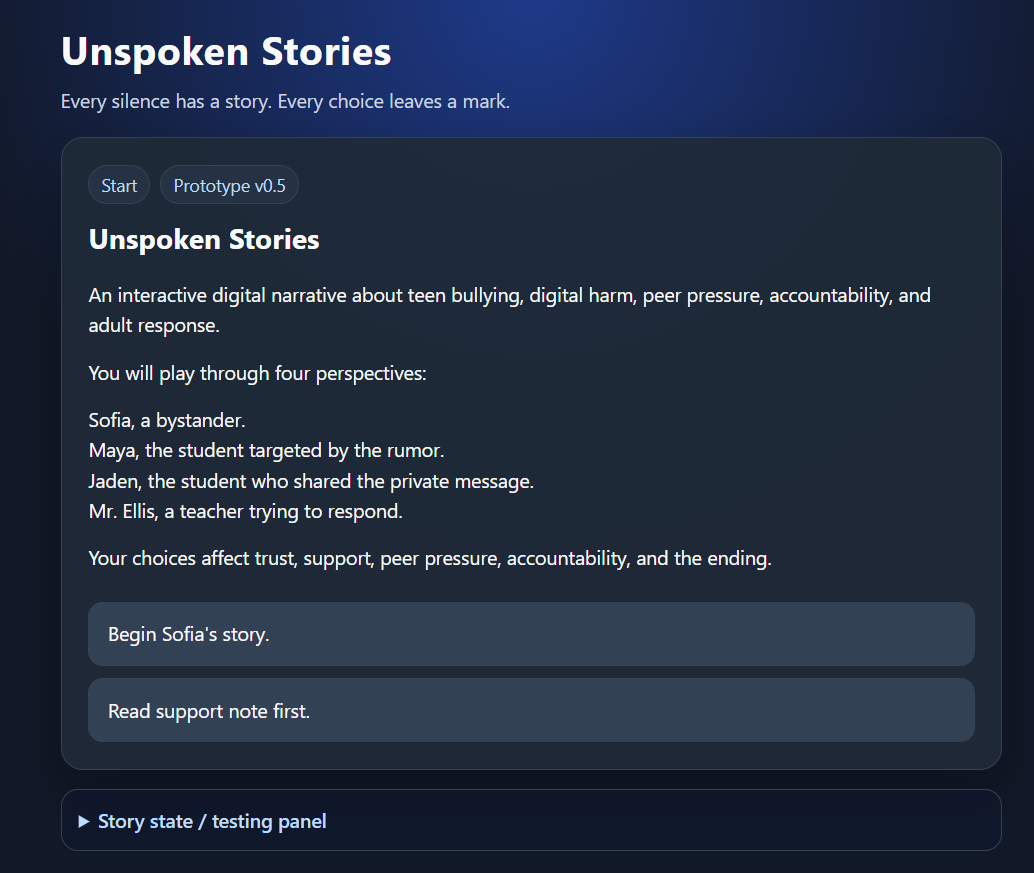}
    \caption{GPT 5.5 generation for T16 Iteration 5}
    \label{fig:GPTT16}
\end{figure}


%
%
\clearpage
 \bibliographystyle{splncs04}
 \bibliography{references}
%


\end{document}